\documentclass[10pt,twocolumn,letterpaper]{article}

\usepackage[pagenumbers]{cvpr} 
\usepackage{array}
\usepackage{makecell}
\usepackage{algorithm}
\usepackage{algpseudocode}
\usepackage{graphicx}
\usepackage{booktabs}
\usepackage{multirow}
\usepackage{wrapfig}
\definecolor{cvprblue}{rgb}{0.21,0.49,0.74}
\usepackage[pagebackref,breaklinks,colorlinks,allcolors=cvprblue]{hyperref}

\def\paperID{16136} 
\def\confName{CVPR}
\def\confYear{2025}

\title{Quantization-Aware Imitation-Learning for Resource-Efficient Robotic Control}

\author{Seongmin Park$^{1}$,    Hyungmin Kim$^{1}$,    Wonseok Jeon$^{2}$,    Juyoung Yang$^{2}$, \\ {Byeongwook Jeon$^{2}$},    {Yoonseon Oh$^{1}$} and {Jungwook Choi$^{1}$}\thanks{Corresponding author}  \\
        \normalsize{\textsuperscript{1}Hanyang University},
        \normalsize{\textsuperscript{2}Hyundai Motor Company} \\
        \normalsize{Seoul, Republic of Korea}\\
        \small{\textsuperscript{1}\texttt{\{skstjdals, kong4274, yoh21\}@hanyang.ac.kr}} \\
        \small{\textsuperscript{2}\texttt{\{wsjeon, yjy6711, smiler\}@hyundai.com}, \textsuperscript{1*}\texttt{choij@hanyang.ac.kr}} \\   
}

\begin{document}
\maketitle

\begin{abstract}

Deep neural network (DNN)-based policy models like vision-language-action (VLA) models are transformative in automating complex decision-making across applications by interpreting multi-modal data. However, scaling these models greatly increases computational costs, which presents challenges in fields like robot manipulation and autonomous driving that require quick, accurate responses. To address the need for deployment on resource-limited hardware, we propose a new quantization framework for IL-based policy models that fine-tunes parameters to enhance robustness against low-bit precision errors during training, thereby maintaining efficiency and reliability under constrained conditions. Our evaluations with representative robot manipulation for 4-bit weight-quantization on a real edge GPU demonstrate that our framework achieves up to 2.5$\times$ speedup and 2.5$\times$ energy savings while preserving accuracy. For 4-bit weight and activation quantized self-driving models, the framework achieves up to 3.7$\times$ speedup and 3.1$\times$ energy saving on a low-end GPU. These results highlight the practical potential of deploying IL-based policy models on resource-constrained devices.


\end{abstract}

\section{Introduction}
\label{sec:introduction}

In recent years, deep neural network (DNN)-based policy models have significantly impacted robot manipulation and autonomous driving~\cite{zhang2021end,kim24openvla,driess2023palm,brohan2022rt,brohan2023rt,liang2018cirl}, primarily by surpassing traditional search-based methods through imitation learning (IL) from expert data. However, these models still struggle with generalization and robustness due to limited data, making it challenging to transfer trained policies across different robot embodiments, tasks, or environments. To address these limitations, there is a growing interest in developing large-scale IL models by adopting foundation models for robotic control~\cite{brohan2022rt, brohan2023rt}.

Vision-Language Action (VLA) models~\cite{kim24openvla,driess2023palm,brohan2022rt,brohan2023rt} combine visual and language capabilities from pretrained vision models, large language models (LLMs)~\cite{gpt4,touvron2023llama2,vicuna2023,geminiteam2023gemini}, and vision-language models (VLMs)~\cite{liu2024visual,alayrac2022flamingo}. Using imitation learning (IL), VLA models learn policies directly from expert demonstrations, equipping them to handle both visual and text information and enhancing their capabilities in robot manipulation tasks. The Open-X-Embodiment project’s RT-X models, trained on large-scale data, improve cross-robot transferability~\cite{o2023open}. While these IL models hold promise as foundation models by supporting cross-embodiment transfer, they are hindered by slow inference speeds, high computational costs, and intensive memory requirements~\cite{wen2024tinyvla}. Since IL models for robotic control often run on resource-constrained, battery-powered devices, these resource demands present significant challenges for efficient implementation.

\begin{figure*}[t]
\centering
{\includegraphics[width=1.75\columnwidth]{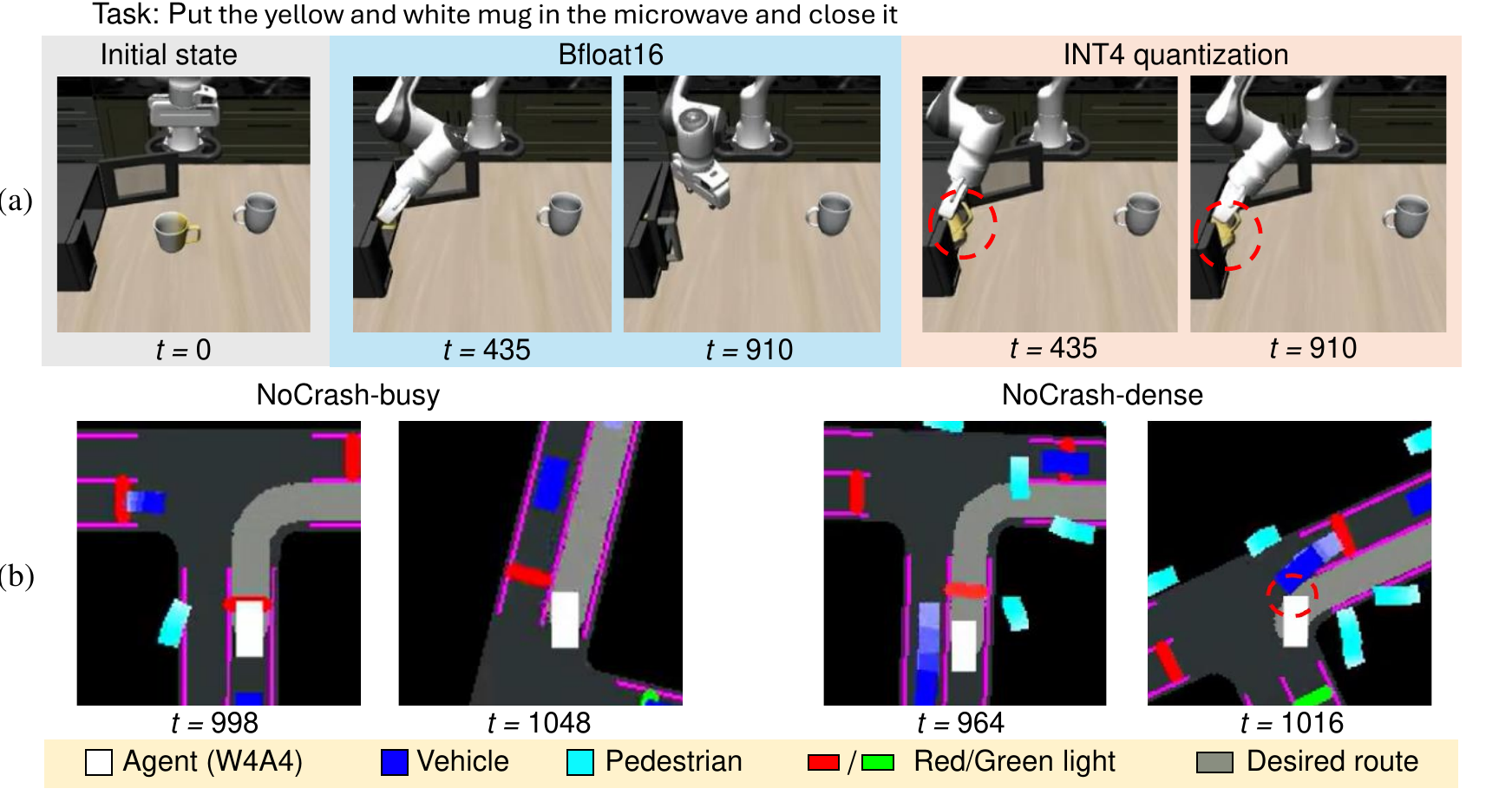}}
\caption{(a) Differences in robot action between INT4 quantization and Bfloat16 in \textit{OpenVLA} on LIBERO. Bfloat16 successfully places the mug inside the microwave and closes the door, whereas INT4 quantization fails in precise action, resulting in the inability to place the mug inside the microwave, leading to task failure. (b) Driving differences of quantized agent \textit{CILRS} (W4A4) at intersections based on benchmark difficulty. (Left) In the relatively easier NoCrash-busy benchmark, the agent drives through intersections without collisions, but (Right) in the more challenging NoCrash-dense benchmark with many pedestrians and vehicles, collisions with other vehicles occur. Note that $t$ represents the timestep.}
\label{fig:bench_difficulty_comparision}
\vspace{-0.2 cm}
\end{figure*}

Quantization offers a promising solution to reduce computational and memory costs by converting neural network weights and activations to lower precision~\cite{choi2018pact,esser2020learned,lin2023awq,frantar2023optq}. However, applying quantization to policy models in applications such as robot manipulation and autonomous driving can lead to performance degradation. Quantization errors directly impact decision-making, often resulting in suboptimal actions. For instance, as shown in Fig.~\ref{fig:bench_difficulty_comparision}(a), complex robot manipulation tasks can suffer due to quantization-induced disturbances in action selection. Similarly, in autonomous driving (Fig.~\ref{fig:bench_difficulty_comparision}(b)), quantization errors can impair steering decisions, increasing the risk of navigation errors and potential collisions. These challenges are critical as such models interact directly with dynamic environments.


We propose a simple yet effective quantization technique for IL-based policy models in robotic control, termed quantization-aware imitation learning (QAIL). This framework integrates quantization into the fine-tuning process, enhancing the policy model’s robustness to quantization errors. Unlike traditional quantization-aware training (QAT) in supervised learning~\cite{hwang2014fixed,zhou2016dorefa,choi2018pact,esser2020learned}, QAIL does not rely on unique labels for each action; instead, it optimizes policy models via IL to maximize the likelihood of action prediction. However, due to a long sequence of actions, quantization errors at each action prediction can accumulate over the sequence, deviating from the full-precision (FP32) model's action distribution. To address this, we introduce quantization-robust behavior cloning (QBC), which encourages the quantized policy to align with the general action selection of the FP32 policy, improving robustness throughout the sequence. This approach enables efficient deployment of quantized policies on resource-constrained devices while preserving accurate decision-making. 
For robot manipulation, evaluations using \textit{OpenVLA}~\cite{kim24openvla} on the LIBERO~\cite{liu2024libero} benchmark show that our 4-bit weight quantized models achieve success rates comparable to FP32 baselines, achieving 2.5× speedup and 2.5× energy savings measured on edge GPU. In autonomous driving, evaluations using \textit{CILRS}~\cite{zhang2021end} on the NoCrash benchmark~\cite{codevilla2019exploring} show that our 4-bit weight and activation quantized self-driving models achieve success rates comparable to FP32 baselines, achieving 3.7× speedup and 3.1× energy savings measured on low-end GPU. Additionally, 8-bit weight and activation quantized models on an on-device CPU achieve 1.7× speedup and 1.3× energy savings.
These results validate our approach as the first to recover and deploy the performance of quantized IL-based policies successfully.


\section{Related Works}
\label{sec:related}

\begin{figure*}[t]

{\includegraphics[width=2.05\columnwidth]{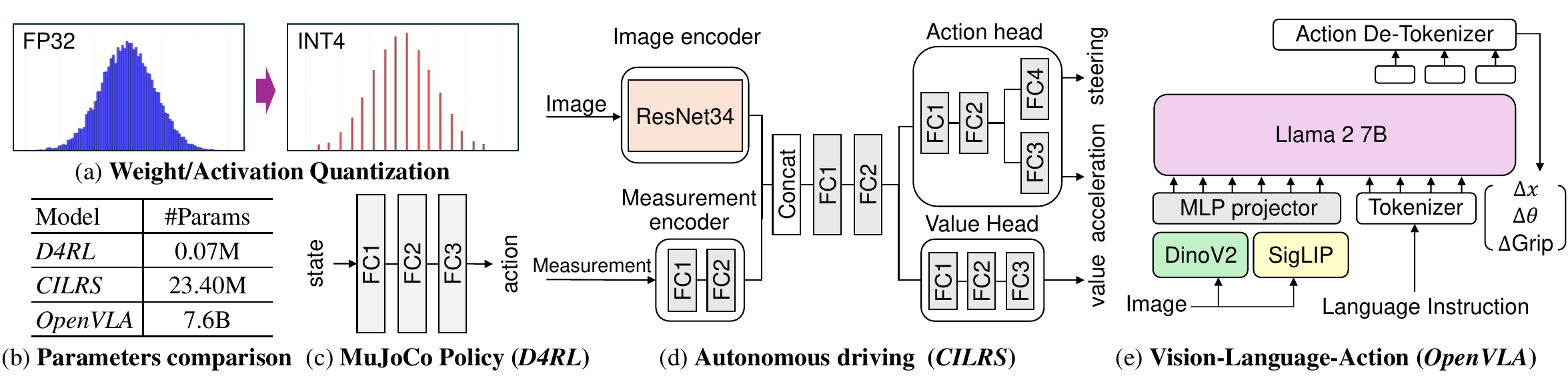}}
\vspace{-0.3 cm}
\caption{Comparison of the structure, number of parameters of DNN-based policy models.}
\label{fig:d4rl_ad_vla}
\vspace{-0.3 cm}
\end{figure*}

\subsection{Quantization for Efficient DNN Inference}
Quantization reduces neural networks' computational and memory requirements by decreasing the bit-precision of weights and/or activations. Quantization is generally applied to the input operands of general matrix multiplication (GEMM), with either one or both operands quantized depending on the bottleneck. For example, convolutional models, which are compute-intensive, benefit from quantizing both weights and activations~\cite{he2016deep,chollet2017xception,simonyan2014very}, as reducing bit precision doubles computation speed (operations per second, OPS)~\cite{markidis2018nvidia} thanks to improved energy efficiency~\cite{horowitz20141}. In contrast, large language models (LLMs), which experience bottlenecks in weight loading times, often employ weight-only quantization strategies~\cite{lin2023awq,frantar2023optq}. In both cases, quantization helps alleviate the bottleneck to enhance inference speed and energy efficiency. 

Fig.~\ref{fig:d4rl_ad_vla}(a) illustrates one of the most widely used quantization techniques, round-to-nearest (RTN)~\cite{jacob2018quantization}. Since continuous-domain values are mapped to a limited set of discrete states, quantization introduces numerical errors that can degrade model accuracy. Research has explored data manipulation techniques to mitigate this degradation~\cite{krishnamoorthi2018quantizing,banner2019post,tambe2020algorithm}. For instance, AWQ~\cite{lin2023awq} reshapes weight distributions to be more compatible with weight-only quantization for LLMs. However, this does not fundamentally enhance the model’s resilience to quantization errors. In contrast, quantization-aware training (QAT) incorporates quantization errors during training, adjusting model parameters to adapt to reduced precision, thereby preserving accuracy~\cite{hwang2014fixed,zhou2016dorefa,choi2018pact,esser2020learned}. For LLMs, where full parameter tuning is prohibitively expensive, low-rank accuracy adapters offer a solution~\cite{dettmers2024qlora,li2023loftq,gunter2024apple}: base model parameters are kept in reduced precision to save memory, while a small set of adapter parameters are fine-tuned to compensate for quantization errors.

\subsection{DNN-based Policy for Imitation Learning}

Deep Neural Network (DNN)-based policy models have driven significant progress across domains like physics simulation, autonomous driving, and robotic manipulation. In physics simulation environments such as MuJoCo~\cite{todorov2012mujoco}, DNN-based models efficiently control a robot’s joints in continuous spaces, trained to solve benchmarks like \textit{D4RL}~\cite{fu2020d4rl}. For autonomous driving, models like \textit{CILRS}~\cite{zhang2021end} process camera images and vehicle data to control acceleration and steering by sampling output distributions to maximize safe driving rewards. In robotic manipulation, vision-language-action models such as RT-2~\cite{brohan2023rt} and \textit{OpenVLA}~\cite{kim24openvla} leverage LLMs to integrate vision and language for multi-task manipulation, predicting action tokens for roll, pitch, yaw, and grasp. These models are trained via imitation learning (IL) to replicate expert behavior efficiently from pre-collected demonstration datasets.

As IL applications expand in scope, DNN-based policy models have increased dramatically in size. Fig.~\ref{fig:d4rl_ad_vla}(c-e) highlights representative IL models for physics simulation (D4RL), self-driving (\textit{CILRS}), and vision-language-action (\textit{OpenVLA}), with model sizes summarized in Fig.~\ref{fig:d4rl_ad_vla}(b). D4RL's basic DNN with three fully connected layers typifies classical physics simulations, while CILRS combines ResNet-based image encoders with fully connected layers for steering an autonomous agent. In contrast, \textit{OpenVLA} integrates the Llama2-7B language model~\cite{touvron2023llama2} with visual features from DINOv2~\cite{oquab2023dinov2} and SigLIP~\cite{zhai2023sigmoid}, forming a robust foundation for versatile robotic manipulation. As shown in Fig.~\ref{fig:d4rl_ad_vla}(b), this expansion has dramatically increased model size by 100,000 times (from 0.07M parameters in D4RL to 7.6B in \textit{OpenVLA}), underscoring the need for efficient inference techniques to reduce computation and memory demands under resource-constrained on-device platforms.

Despite increasing demands for efficient policy inference, few attempts have been made, primarily within reinforcement learning (RL). Early work, such as low-precision policy distillation (LPPD~\cite{mckinstry2018low}), applied quantization-aware training (QAT) with aggressive bit reduction (1-bit) to a classical policy model. Quantization in RL models, focusing on the policy network~\cite{faust2022quarl,zhang2023fastact}, accelerates training and inference by streamlining interactions of network learners and actors with the environment, improving efficiency while preserving performance. As a complementary approach, structured pruning~\cite{park2024pruning} reduces computational demands without compromising decision-making. However, these methods generally utilize simple RL models and environments, limiting their relevance to more advanced robot control tasks. For applications like autonomous driving and multi-modal action prediction in robotic manipulation, policy models must reliably handle long-tail scenarios with high stability, even under reduced precision.

\section{Background}
\label{sec:background}

\subsection{Imitation Learning}

Imitation Learning (IL) enables an agent to learn policies directly from expert demonstrations, making it ideal for applications where defining a reward function is challenging, such as in robotics and autonomous driving. In this setting, \(S\) represents the set of possible states, and \(A\) the set of possible actions. Given a dataset \(D_E = \{\tau_1, \ldots, \tau_N\}\) with \(N\) expert demonstrations, each demonstration \(\tau_i\) consists of a sequence of state-action pairs of length \(T_i\), denoted as \(\tau_i = \{(s_1, a_1), \ldots, (s_{T_i}, a_{T_i})\}\), where \(s \in S\) and \(a \in A\). These demonstrations are generated by sampling actions from an expert policy \(\pi_E\) under environment dynamics \(\rho(s'|s, a)\), where \(s'\) is the resulting state after action \(a\) is taken in state \(s\).

The goal of IL is to learn a policy \(\pi_\theta: S \to A\) that closely replicates expert behavior. This is often achieved through \textit{behavior cloning}, which optimizes a supervised learning objective to maximize the likelihood of expert state-action pairs in the dataset. The objective loss function is:
\begin{equation}
\label{eq:il_objective}
\mathcal{L}^{IL}(\theta) = -\frac{1}{|\mathcal{D}_E|} \sum_{(s, a) \in \mathcal{D}_E} \log \pi_\theta(a | s).
\end{equation}
Minimizing this loss enables the learned policy \(\pi_\theta\) to imitate the expert by predicting expert actions \(a\) for given states \(s\).

\subsection{Challenges for Policy Quantization}

Policy models like autonomous driving and robotic manipulation achieve their goals through action sequences. However, quantization errors can cause these actions to diverge from those of full-precision (FP) models for a given state, impairing performance, especially in complex, long-tail tasks. For instance, as shown in Fig.~\ref{fig:bench_difficulty_comparision}(a), robotic manipulation tasks that involve interacting with surrounding objects, such as ``put the mug in the microwave and close it,'' are more intricate than simply moving an object to a specific location. Significant quantization errors often lead to failure in this intricate robot manipulation case.

In autonomous driving, Fig.~\ref{fig:bench_difficulty_comparision}(b) illustrates a scenario where a vehicle must execute a safe right turn at an intersection with traffic signals. While a quantized policy can perform well in the more straightforward NoCrash-Busy scenario with minimal pedestrians and vehicles, it struggles in the more challenging NoCrash-Dense scenario, where numerous pedestrians and vehicles increase the risk of collisions. These performance issues are especially pronounced in previously unseen maps, underscoring the difficulty of generalizing quantized models to new environments.

These examples show that quantization errors directly affect action accuracy, limiting the use of quantized policies in mission-critical tasks. Therefore, developing methods that minimize quantization errors is essential to ensure reliable performance in complex, dynamic environments.

 \begin{figure}[t]
\centering
{\includegraphics[width=0.9\columnwidth]{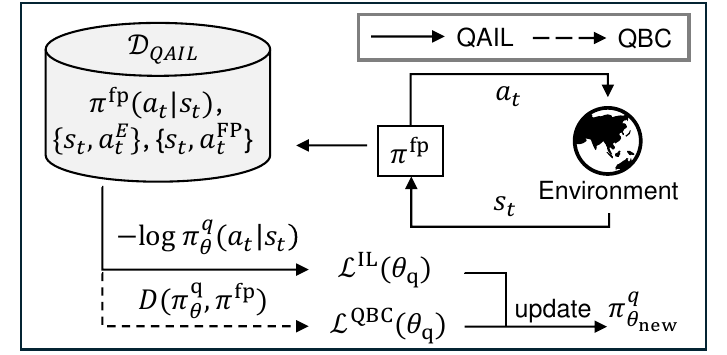}}
\vspace{-0.3 cm}
\caption{Overview of QAIL+QBC }
\label{fig:qail}
\vspace{-0.4 cm}
\end{figure}

\section{Method}
\label{sec:method}

\subsection{Quantization-Aware Imitation Learning}

Quantization-Aware Imitation Learning (QAIL) integrates quantization into IL, enabling low-precision policies that enhance efficiency and reduce memory usage, particularly suited for resource-constrained devices. As shown in Fig.~\ref{fig:qail}, when an interactive environment is available, the combined dataset \(\mathcal{D}_{\text{QAIL}}\) merges the expert demonstration dataset \(\mathcal{D}_{\text{E}}\) with additional state-action pairs \(\mathcal{D}_{\text{FP}}\) from the full-precision (FP) policy \(\pi^{\text{FP}}\), allowing the quantized policy \(\pi^{\text{q}}\) to be trained via imitation learning. The QAIL loss function, defined as:
\begin{equation}
\label{eq:qail_objective}
\mathcal{L}^{\text{QAIL}}(\theta) = - \mathbb{E}_{(s_t, a_t) \sim \mathcal{D}_{\text{QAIL}}} \left[ \log \pi^{\text{q}}_{\theta}(a_t|s_t) \right],
\end{equation}
enables \(\pi^{\text{q}}\) to approximate expert behavior similar to \(\pi^{\text{FP}}\). While quantization-aware training (QAT) performs well in perception tasks~\cite{choi2018pact,esser2020learned} where errors affect only within a single inference instance, in sequential tasks like autonomous driving and robotic manipulation, even small action errors can accumulate over a sequence, making these tasks highly sensitive to quantization errors. Thus, directly combining IL and QAT, as in QAIL, often fails to maintain full-precision model performance in these domains.

\begin{algorithm}[t]
\caption{Quantization-Aware Imitation Learning}
\begin{algorithmic}
\State \textbf{Input:} Pre-trained policy $\pi^{\text{fp}}$, quantized policy $\pi^{\text{q}}_{\theta}$, hyperparameter $\lambda$, Expert dataset $\mathcal{D}_{\text{E}}$
\State \textbf{Output:} Updated policy $\pi^{\text{q}}_{\theta}$

\State Initialize $\pi^{\text{q}}_{\theta}$ from $\pi^{\text{fp}}$

\State FP policy dataset $\mathcal{D}_{\text{FP}} = \emptyset$
\State Collect state-action pairs using $\pi^{\text{fp}}$ and populate $\mathcal{D}_{\text{FP}}$
\State Combine datasets: $\mathcal{D}_{\text{QAIL}} = \mathcal{D}_{\text{FP}} \cup \mathcal{D}_{\text{E}}$
\For{each state-action pair $(s_t, a_t)$ in $\mathcal{D}_{\text{QAIL}}$}
    \State Calculate $\mathcal{L}^{\text{total}} = \mathcal{L}^{\text{IL}} + \lambda \mathcal{L}^{\text{QBC}}$ using Eq.~(\ref{eq:qail_objective}) and (\ref{eq:qbc_objective})
    \State Compute the gradient $\frac{\partial \mathcal{L}^{\text{total}}}{\partial \theta} = \frac{\partial \mathcal{L}^{\text{total}}}{\partial \theta_{\text{q}}} \cdot \frac{\partial \theta_{\text{q}}}{\partial \theta} \underset{\text{STE}}{\approx} \frac{\partial \mathcal{L}^{\text{total}}}{\partial \theta_{\text{q}}}$
    \State Update $\pi^{\text{q}}_{\theta}$ by minimizing $\mathcal{L}^{\text{total}}$
\EndFor

\State \textbf{return} Updated policy $\pi^{\text{q}}_{\theta}$
\end{algorithmic}
\label{algo:alg2}
\end{algorithm}


\setlength{\columnsep}{15pt}
\begin{wrapfigure}{r}{0.2\textwidth}
\vspace{-0.4 cm}
    \centering
    \includegraphics[width=\linewidth]{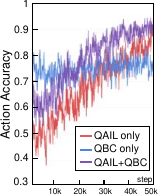}
   \caption{Action accuracy Comparison.}
\label{fig:action_acc}
\vspace{-0.5 cm}
\end{wrapfigure}
\begin{figure*}
    \centering
    \includegraphics[width=0.8\linewidth]{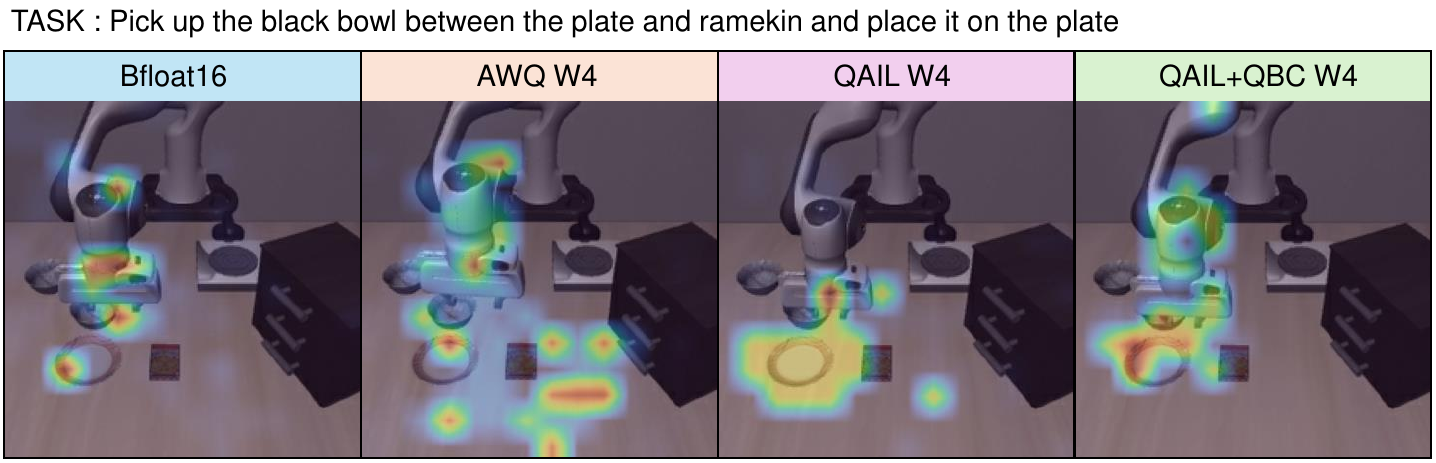}
\vspace{-0.1 cm}
   \caption{Comparison of attention visualization for tasks successfully completed on the LIBERO-Spatial benchmark. Additional examples are provided in ~\ref{sec:appendix_map}.}
\vspace{-0.2 cm}
\label{fig:attention}
\end{figure*}
\subsection{Quantization-Robust Behavior Cloning}
\label{sec:qbc}
To address QAIL's challenge, we propose quantization-robust behavior cloning (QBC), leveraging the FP policy alongside demonstration data to minimize quantization errors by aligning the action distributions of the quantized and FP policies. The QBC loss function measures the discrepancy between the quantized and FP policies’ action distributions, defined as:
\begin{equation}
\label{eq:qbc_objective}
\mathcal{L}^{\text{QBC}}(\theta) = \mathbb{E}_{s_t \sim \pi^{\text{q}}_{\theta}} \left[ D(\pi^{\text{q}}_{\theta}(a_t|s_t), \pi^{\text{FP}}(a_t|s_t)) \right],
\end{equation}
where \(D\) is a discrepancy metric (e.g., L2-norm). The total loss function combines QAIL and QBC losses:
\begin{equation}
\label{eq:total_objective}
\mathcal{L}^{\text{total}}(\theta)  = \mathcal{L}^{\text{QAIL}}(\theta)+\lambda\mathcal{L}^{\text{QBC}}(\theta),
\end{equation}
with \(\lambda\) balancing QBC’s influence. This combined approach effectively reduces quantization errors, enabling the quantized policy to generalize as effectively as the FP policy. Fig.~\ref{fig:qail} and Algorithm~\ref{algo:alg2} outline the full procedure, and Fig.~\ref{fig:action_acc} illustrates that QAIL+QBC consistently achieves higher action accuracy than either method alone, demonstrating the synergy between the two loss components.

We propose weighted QBC (wQBC) to enhance performance in long-horizon tasks where quantization errors can markedly accumulate over time. This accumulation necessitates more refined methods to mitigate the impact of quantization errors over extended action sequences. While imitation learning aims to mimic expert demonstrations, exact replication isn't always ideal due to multiple acceptable actions for a given state. Overemphasizing precise imitation can lead to overfitting and reduce generalization.

To address this, we adjust the influence of action cloning based on state importance, allowing the policy to focus on critical decisions that affect long-term success. We measure state importance using a saliency score inspired by perturbation-based visual attention mechanisms~\cite{greydanus2018visualizing}: 
\begin{equation}
    S_{\pi}(i,j) = \frac{1}{2} \left\| \pi(I) - \pi(\phi(I,i,j)) \right\|^2,
\end{equation}
\begin{equation}
\overline{S_{\pi}} = \frac{1}{I \times J} \sum_{i=1}^I \sum_{j=1}^J S_{\pi}(i,j),
\\
\end{equation}
where $\phi(I,i,j)$ applies a Gaussian filter at location $(i,j)$ to perturb the image, assessing how input alterations influence the policy's decisions.
Building on this, we define the wQBC loss function to selectively enhance learning: 
\begin{equation}
\label{eq:dataset_qdpo_roach}
\mathcal{L}^{\text{wQBC}} = \alpha \cdot \mathcal{L}^{\text{QBC}},  \text{where } \alpha = \begin{cases} 
\beta & \text{if } \overline{S_{\pi^{\text{q}}}} > T, \\
1 & \text{otherwise,}
\end{cases}
\end{equation}



\setlength{\columnsep}{12pt}
\begin{wraptable}{r}{0.5\columnwidth}
\vspace{-0 cm}
\centering
\resizebox{0.5\columnwidth}{!}{%
\begin{tabular}{l|c}
\Xhline{2\arrayrulewidth}
Method & LIBERO-Long \\ \hline
QAIL+QBC & 47.8 \% \\ \midrule
QAIL+wQBC & \textbf{50.4\%} \\  \Xhline{2\arrayrulewidth}
\end{tabular}
}
\vspace{-0.2 cm}
\caption{Success rate comparison of \textit{OpenVLA} with QBC and wQBC.}
\vspace{-0.3 cm}
\label{tab:wqbc}
\end{wraptable}
where, \(\beta\) is a hyperparameter greater than 1. During implementation, the threshold \(T\) is set to differentiate the top 20\% of saliency scores, calculated from 10\% of the fine-tuning dataset. As demonstrated in Table~\ref{tab:wqbc}, wQBC shows improved performance on long-horizon tasks compared to standard QBC by effectively focusing the policy's learning on the most influential states.

\begin{table}
\centering
\resizebox{0.9\columnwidth}{!}{%
\begin{tabular}{l|c|c|c|c}
\Xhline{2\arrayrulewidth}
\multirow{3}{*}{Method} & \multicolumn{4}{c}{AttDiv $\downarrow$} \\ \cline{2-5} 
& \makecell{LIBERO\\-Spatial} & \makecell{LIBERO\\-Object} & \makecell{LIBERO\\-Goal} & \makecell{LIBERO\\-Long}\\ \midrule
AWQ & 0.0821 & 0.0268 & 0.0612 &  0.0640\\ \midrule
QAIL + QBC & 0.0559& 0.0198 & 0.0458  &  0.0521\\ \Xhline{2\arrayrulewidth}
\end{tabular}
}
\vspace{-0.2 cm}
\caption{Comparison of AttDiv with \textit{OpenVLA} on the LIBERO.}
\vspace{-0.3 cm}
\label{tab:attsim}
\end{table}

\subsection{Analysis}
\label{sec:analysis}
\textbf{Attention Visualization}.
We visualize attention to understand the behavioral differences between the quantized policy and FP policy. Additionally, we explore how QAIL+QBC modifies the quantized policy. We employ a perturbation-based attention~\cite{greydanus2018visualizing} to extract attention maps for the policy model. 

As depicted in Fig.~\ref{fig:attention}, the FP policy shows high attention on target objects or target placements and locations where the robot arm and objects are present. However, policies quantized using PTQ (AWQ) often attend to areas where objects are absent or not important. QAIL demonstrates a notable improvement by focusing more around objects. However, it still shows a tendency to attend to regions where no objects are present. In contrast, applying QAIL+QBC for fine-tuning directs attention closely to important locations similar to the FP policy, confirming that this approach enables the quantized policy to perform actions through reasoning akin to the FP policy.

Furthermore, we measure the attention divergence (AttDiv) between the quantized and FP models to quantify their attention similarity, calculated as follows:
\begin{equation}
    \text{AttDiv}(\pi^{\text{q}}, \pi^{\text{fp}}) = D_{\text{KL}}(S_{\pi^{\text{q}}}\parallel S_{\pi^{\text{fp}}})
\end{equation}
Table~\ref{tab:attsim} demonstrates that the attention divergence in QAIL+QBC consistently lower compared to the AWQ.

\begin{table*}
\centering
\resizebox{1.7\columnwidth}{!}{%
\begin{tabular}{l|c|c|c|c|c|c}
\Xhline{2\arrayrulewidth}
\multirow{2}{*}{Method} & \multirow{2}{*}{Bit-width} & \multicolumn{5}{c}{Success Rate $\uparrow$} \\ \cline{3-7} 
 & & LIBERO-Spatial & LIBERO-Object & LIBERO-Goal & LIBERO-Long & Average \\ \midrule
Baseline & Bfloat16 & 83.6\% & 83.8\% & 76.6\% & 50.8\% & 74.0\% \\ \midrule
AWQ & INT4 & 80.0\% & 81.2\% & 74.6\% & 47.2\% & 70.8\% \\ 
QAIL & INT4 & 81.0\% & 82.2\% & 75.8\% & 47.0\% & 71.5\% \\ 
QAIL + QBC & INT4 & \textbf{84.4\%} & \textbf{83.6\%} & \textbf{76.4\%} & \textbf{*50.4\%} & \textbf{73.1\%} \\ \Xhline{2\arrayrulewidth}
\end{tabular}
}
\caption{Comparison of success rate across various quantization methods with \textit{OpenVLA} and scenarios in the LIBERO benchmark.\\ * wQBC was specifically applied in the LIBERO-Long.}
\vspace{-0.1 cm}
\label{tab:libero_comparison}
\end{table*}

\begin{figure}[t]
\centering
{\includegraphics[width=0.85\columnwidth]{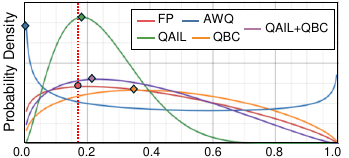}}
\caption{Comparison of action distributions at a specific state \(s_j\) for different quantization methods applied to \textit{CILRS} (W4A4) on the NoCrash-dense benchmark. The red circle represents the action \(a_j\) from the demonstration dataset $\mathcal{D}_{\text{FP}}$ at state \(s_j\), while the diamond shape indicates the point of maximum probability for each distribution.}
\vspace{-0.2 cm}
\label{fig:act_dist}
\end{figure}

\textbf{Action distribution comparison.} We compare the action distributions within the \textit{CILRS} using different quantization methods: PTQ (RTN~\cite{jacob2018quantization}), QAIL, QBC, QAIL+QBC. This model outputs a probability distribution from which actions are sampled. Our analysis focuses on the action distributions for a specific state to illustrate how different quantization methods impact the policy's behavior.

As shown in Fig.~\ref{fig:act_dist}, PTQ introduces significant quantization errors, resulting in an action distribution that markedly differs from the FP policy, failing to capture its intended behavior. QAIL trains the quantized policy to increase the likelihood of actions from the demonstration dataset, leading to distributions with excessively high peaks for those specific actions. This skews the original distribution of the FP policy, especially at the peaks. On the other hand, QBC aims to align the quantized policy’s action distribution with that of the FP policy by minimizing their discrepancy. However, because QBC does not use actions from the demonstration data, it can miss high-quality expert actions. When the FP policy’s distribution is broad or nearly uniform, indicating multiple acceptable actions, a policy trained with QBC alone might select statistically similar but suboptimal actions, reducing decision quality.

By combining QAIL and QBC, we leverage their strengths while mitigating individual limitations. This approach trains the quantized policy to both mirror the FP policy's action distribution and increase the likelihood of high-quality actions from the demonstrations. Consequently, QAIL+QBC produces an action distribution closely aligned with both the FP policy and expert demonstrations, overcoming the weaknesses of using either method alone:

\begin{itemize} \item QAIL alone emphasizes high-quality actions but may distort the action distribution due to overfitting. \item QBC alone maintains structural similarity to the FP policy's distribution but may overlook optimal actions from demonstrations. \end{itemize}
Integrating both methods enables the quantized policy to retain the FP policy's overall behavior while prioritizing expert-recommended actions, leading to higher-quality decisions and improved performance.

\begin{table*}[t]
\centering
\resizebox{0.85\textwidth}{!}{%
\begin{tabular}{l|c|c|c|c|c|c|c|c|c|c|c|c}
\Xhline{2\arrayrulewidth}
\multirow{2}{*}{Method} & \multirow{2}{*}{Bit-width} & \multicolumn{11}{c}{Success Rate \% $\uparrow$} \\ \cline{3-13}
 & & task1 & task2 & task3 & task4 & task5 & task6 & task7 & task8 & task9 & task10 & Avg \\ \midrule
baseline & Bfloat16 & 86 & 92 & 82 & 98 & 72 & 88 & 92 & 82 & 82 & 62 & 83.6 \\ \midrule
 RTN& INT4& 82 & 66 & 74 & 88 & 52 & 82 & 92 & 76 & 62 & 58 & 73.2 \\ 
 AWQ& INT4& 84 & 84 & 72 & 80 & 74 & 88 & \textbf{94} & \textbf{80} & 76 & 68 & 80.0 \\ 
 QBC only& INT4& 74 & 78 & 74 & 88 & 64 & 76 & 76 & 66 & 46 & 52 & 69.0 \\ 
 QAIL only& INT4& \textbf{86} & 90 & 78 & 78 & 70 & \textbf{94} & 92 & 74 & 72 & 76 & 81.0 \\ 
 QAIL+QBC& INT4& \textbf{86} & \textbf{94} & \textbf{86} & \textbf{92} & \textbf{76} & 88 & \textbf{94} & 70 & \textbf{78} & \textbf{80} & \textbf{84.4} \\ \Xhline{2\arrayrulewidth}
\end{tabular}
}
\vspace{-0.2 cm}
\caption{Comparison of success rates across different tasks in the LIBERO-Spatial benchmark.}
\vspace{-0.2 cm}
\label{tab:libero_ablation}
\end{table*}

\begin{table*}[h!]
\centering
\resizebox{1.55\columnwidth}{!}{%
\begin{tabular}{l|c|cccc|cccc|cccc}
\Xhline{2\arrayrulewidth}
\multirow{2}{*}{Method} & \multirow{2}{*}{Bit-width} & \multicolumn{4}{c|}{Suc. Rate \% $\uparrow$} & \multicolumn{4}{c|}{Dri. Score \% $\uparrow$} & \multicolumn{4}{c}{Reward $\uparrow$} \\ \cline{3-14} 
 &  & tt & tn & nt & nn & tt & tn & nt & nn & tt & tn & nt & nn \\ \Xhline{2\arrayrulewidth}
Baseline & FP & 82 & 74 & 80 & 68 & 78 & 71 & 80 & 64 & 2180 & 1988 & 1223 & 1343 \\ \midrule
RTN & W4A4 & 34 & 43 & 36  & 29  & 30  & 38 & 35  & 31 & 324  & 348  & 289  & 195  \\
QAIL & W4A4 & 62 & 58 & 58 & 48 & 62 & 62 & 58 & 58 & 883 & 745 & 540 & 529 \\ 
QAIL+QBC & W4A4 & \textbf{80} & \textbf{70} & \textbf{74} & \textbf{70} & \textbf{77} & \textbf{70} & \textbf{76} & \textbf{62} & \textbf{2094} & \textbf{1903} & \textbf{1001} & \textbf{1005} \\ \Xhline{2\arrayrulewidth}
\end{tabular}%
}
\caption{Comparison of success rate, driving score, and reward for different quantization methods with \textit{CILRS} on the NoCrash-dense benchmark. (tt: train-town \& train-weather, tn: train-town \& new-weather, nt: new-town \& train-weather, nn: new-town \& new-weather)}
\label{tab:cilrs_main}
\end{table*}

\section{Experiments}
\label{sec:experiments}
\subsection{Experimental Settings}

We conducted experiments on the quantization of DNN-based policy models across various domains, including robot manipulation, autonomous driving, and classical physics simulation. Detailed experimental settings for each task are provided in \ref{sec:appendix_exp}.

\textbf{Robot Manipulation:} We employed the \textit{OpenVLA}~\cite{kim24openvla} and applied weight quantization using the AWQ~\cite{lin2023awq} method for QAIL. Evaluation was carried out on simulated robot setups and tasks using the LIBERO~\cite{liu2024libero} benchmark, which assesses models' understanding of spatial relationships and object types across different modules: LIBERO-Spatial, LIBERO-Object, and task-oriented behavior in LIBERO-Goal, as well as long-horizon tasks in LIBERO-Long. Each target dataset was fine-tuned using QAIL+QBC, and we trained a minimal set of trainable parameters, totaling 110M, using QLoRA (r=32). For each task, a pretrained model served as the FP policy.

\textbf{Autonomous Driving:} We utilized the \textit{CILRS}~\cite{zhang2021end} and collected data using the CARLA~\cite{dosovitskiy2017carla} simulator. Self-driving model being compute bound, required both weight and activation quantization to achieve latency improvements; thus, activation quantization was also performed. For PTQ quantization, we applied RTN~\cite{jacob2018quantization}, and for QAIL, we utilized LSQ~\cite{esser2020learned}. Both techniques were employed to quantize the models to 4-bit precision.
The models were evaluated on the NoCrash-dense benchmark~\cite{codevilla2019exploring}, which specifies training and evaluation conditions across different towns and weather scenarios. 

\textbf{Classical Physics Simulation:} Fine-tuning and evaluation were conducted using the D4RL benchmark~\cite{fu2020d4rl}, where both weights and activations were quantized to 4-bit precision using the LSQ.

\subsection{Experimental Results}

\subsubsection{Robot Manipulation}

Table~\ref{tab:libero_comparison} presents a comparison of performance on the LIBERO benchmark across different quantization techniques.  Specifically, when only AWQ is applied, it results in a 3.2\% performance reduction compared to the baseline. This outcome is particularly concerning for robot manipulation tasks, where safety is closely tied to the interaction with the real world. In contrast, QAT approaches like QAIL show an improved average success rate, increasing by 0.7\%. However, this is still below the performance of the baseline model using Bfloat16, which achieves 74.0\% success rate. Using QAIL+QBC, leveraging additional information from the FP policy allows for a narrower performance gap with the FP model. This enhanced approach shows a 2.3\% increase in performance over AWQ, demonstrating its potential in resource-constrained environments.

To evaluate the individual effects of different quantization methods and the unique contributions of QBC and QAIL, we evaluate each approach separately, as detailed in Table~\ref{tab:libero_ablation}. Applying basic RTN quantization without any calibration for reducing quantization errors results in a significant performance degradation of 13.4\% compared to the baseline. Using QAIL alone, which utilizes the expert dataset as a ground truth for fine-tuning, shows performance improvements over AWQ. However, applying QBC alone, which aims to reduce discrepancies with the FP policy without clear ground truth, actually results in a decrease in performance. In contrast, the combination of QBC and QAIL losses brings substantial performance improvements, as QBC acts as a positive guide. 

\begin{table}[t]
    \centering
    \resizebox{0.95\columnwidth}{!}{%
    \begin{tabular}{l|c|c|c|c}
    \Xhline{2\arrayrulewidth}
    \multirow{3}{*}{Method} & \multirow{3}{*}{Bit-width} & \multicolumn{3}{c}{Metrics $\downarrow$} \\ \cline{3-5}
     &   & Collision & Collision & Red light \\ 
     &   & pedestrian & vehicle & infraction \\ \Xhline{2\arrayrulewidth}
    IL & FP  & 0.06 & 0.89 & 1.86 \\ \midrule
    QAIL & W4A4  & 0.74 & 1.53 & 3.24 \\ 
    QAIL+QBC & W4A4  & \textbf{0.05} & \textbf{0.93} & \textbf{2.10} \\ \Xhline{2\arrayrulewidth}
    \end{tabular}%
    }
    \caption{Infraction analysis in a new-town and new-weather scenario with \textit{CILRS} on the NoCrash-dense benchmark.}
     \vspace{-0.3 cm}
    \label{tab:cilrs_collision}
\end{table}

\subsubsection{Autonomous Driving}

As seen in Table~\ref{tab:cilrs_main}, the performance comparison after applying quantization methods shows a significant drop in performance in both train and new-towns under the W4A4 setting. However, by applying QBC, the alignment with the driving capabilities of the FP policy is significantly improved, achieving much better performance. Similarly, the infraction analysis in Table~\ref{tab:cilrs_collision} shows that after applying QBC, pedestrian collisions are reduced, demonstrating improved safety.

\subsubsection{Physics Simulation Tasks}
To evaluate the generalization capability of our algorithms, we expand our evaluation from the robot manipulation and Autonomous driving to classical continuous control tasks from the DeepMind Control Suite within D4RL~\cite{fu2020d4rl}, as shown in Table~\ref{tab:mujoco}. This allowed us to compare our quantization techniques QAIL+QBC, against established methods like LPPD~\cite{mckinstry2018low}. The policy network has three fully connected layers with 2048 units each, while the critic network features three layers of 512 units.

\begin{table}[t]
\centering
\resizebox{0.81\columnwidth}{!}{%
\begin{tabular}{l|c||c|c}
\Xhline{2\arrayrulewidth}
\multirow{2}{*}{TASK} & FP32 & \multicolumn{2}{c}{INT4} \\ \cline{2-4} 
 & IL & LPPD\cite{mckinstry2018low} & QAIL+QBC \\ \hline
Cartpole Balance & 652 & 424 & \textbf{635} \\ \hline
Walker Stand & 692 & 442 & \textbf{688} \\ \hline
Hopper Stand & 645 & 404 & \textbf{643} \\ \hline
Cheetah Run &  567 & 294 & \textbf{556} \\ \hline
Finger Spin & 684 & 421 & \textbf{640} \\ \hline
Humanoid Stand & 565 & 356 & \textbf{590} \\ \hline
Humanoid Walk & 550 & 205 & \textbf{535} \\ \Xhline{2\arrayrulewidth}
\end{tabular}%
}
\vspace{-0.2 cm}
\caption{Comparison of average return for each task by quantization method on physics simulation.}
\vspace{-0.2 cm}
\label{tab:mujoco}
\end{table}




\subsection{Ablation study}
\textbf{ViT vs. LLM: Quantization Impact on Performance.} 
\setlength{\columnsep}{12pt}
\begin{wraptable}{r}{0.5\columnwidth}
\vspace{-0.4 cm}
\centering
\resizebox{0.5\columnwidth}{!}{%
\begin{tabular}{c|c|c}
\Xhline{2\arrayrulewidth}
\multicolumn{2}{c|}{\textit{OpenVLA}} & \multirow{2}{*}{Suc. Rate $\uparrow$}\\ \cline{1-2} 
ViT& LLM & \\ \midrule
Bfloat16 & Bfloat16  &  74.0\%  \\ \midrule
INT4 & Bfloat16  & \textbf{73.4\%} \\
Bfloat16 & INT4  & 71.3\% \\
INT4 & INT4  &70.8\% \\ \Xhline{2\arrayrulewidth}
\end{tabular}
}
\vspace{-0.3 cm}
\caption{Quantization impact of ViT and LLM components on the LIBERO for \textit{OpenVLA}.}
\vspace{-0.3 cm}
\label{tab:vit_llm}
\end{wraptable}
As demonstrated in Table~\ref{tab:vit_llm}, the quantization of the ViT component results in minimal performance changes, whereas quantization of the LLM component significantly impacts performance. This suggests that the reasoning capabilities of the LLM are crucial for achieving action decisions similar to those of the FP policy, indicating that QAIL+QBC fine-tuning effectively compensates for the losses from quantizing the LLM. 

\textbf{Reinforcement Learning with QBC.} We also explored the application of QBC in reinforcement learning settings to further investigate its utility in training policies under quantization constraints. The detailed methodology and results are provided in \ref{sec:appendix_qarl}. These experiments further confirm QBC's broad applicability across learning paradigms, highlighting its role in addressing policy quantization challenges.

\subsection{Implementation}

We evaluate the versatility of our quantization techniques by implementing them on various resource-constrained platforms, including an edge device (NVIDIA Jetson AGX Orin) and a low-end GPU (NVIDIA 2080Ti). For practical deployment, we test the autonomous driving models: the W8A8 model on a CPU (ARM Cortex-A78AE) and the W4A4 model on the NVIDIA 2080Ti GPU. Additionally, 8-bit and 4-bit weight-quantized VLA models are implemented on an Jetson AGX Orin's GPU. Latency is measured using 1,000 input samples, while CPU and GPU energy consumption are recorded with tegrastats~\cite{tegrastats_toolkit}. Detailed experimental settings, latency breakdowns, and further analyses are provided in \ref{sec:appendix_imp}.

\textbf{Autonomous Driving:} We employ ncnn~\cite{ni2017ncnn} for W8A8 support with ARM NEON intrinsic and AutoTVM~\cite{chen2018tvm} for W8A8 and W4A4 GPU kernels. As shown in Table~\ref{tab:CILRS_OpenVLA_Performance}, the \textit{CILRS} (CPU) W8A8 model achieves a 1.7$\times$ reduction in latency compared to FP32, along with a 75\% reduction in memory usage and a 30\% improvement in energy efficiency. For the \textit{CILRS} (GPU), the W4A4 model achieves up to 3.7$\times$ lower latency and 3.1$\times$ better energy efficiency.

\textbf{Robot Manipulation:} For GPU implementation of weight-only quantized \textit{OpenVLA}, we use TensorRT for the vision encoder and MLC-LLM~\cite{mlc-llm} for the LLM backbone. Compressed weights mitigate the memory-bound nature of LLM operations, significantly accelerating inference. As shown in Table~\ref{tab:OpenVLA_Performance}, the INT8 model achieves a 1.6$\times$ speedup, while the INT4 model achieves a 2.5$\times$ speedup compared to BFloat16. Energy efficiency improves by 1.7$\times$ with INT8 and 2.5$\times$ with INT4. Additionally, the 4$\times$ memory reduction with INT4 is particularly advantageous for foundation models, which are inherently memory-intensive, making this approach suitable for edge devices with limited resources.

Our evaluations on edge devices demonstrate that the optimized models significantly improve performance and energy efficiency on both CPU and GPU platforms. These results validate that our quantization techniques enable resource-efficient robot control in computationally constrained environments.

\begin{table}[t]
\centering
\resizebox{1\columnwidth}{!}{%
\begin{tabular}{l|c|c|c|c|c|c}
\Xhline{2\arrayrulewidth}
\multirow{2}{*}{Measurement} & \multicolumn{3}{c|}{CPU (Jetson AGX Orin)} & \multicolumn{3}{c}{GPU (RTX 2080Ti)} \\ \cline{2-7}
 & FP32 & FP16 & W8A8 & FP32 & W8A8 & W4A4 \\ \midrule
Memory (MB) & 78.4 & 39.2 & 19.6 & 78.4 & 39.2 & 19.6 \\ 
Latency (ms) & 212.2 & 176.4 & 124.7 & 11.6 & 4.1 & 3.1 \\ 
SpeedUp & $1.0 \times$ & $1.2\times$ & $1.7 \times$ & $1.0 \times$ & $2.9 \times$ & $3.7 \times$ \\ 
Energy Saving & $1.0 \times$ & $1.1 \times$ & $1.3 \times$ & $1.0 \times$ & $2.5 \times$ & $3.1 \times$ \\ \Xhline{2\arrayrulewidth}
\end{tabular}
}
\vspace{-0.2 cm}
 \caption{Performance Metrics for \textit{CILRS} (14.5 GFLOPs) measured on a CPU and a GPU, demonstrating memory consumption, latency and energy saving across the different precision formats.}
    \label{tab:CILRS_OpenVLA_Performance}
\end{table}

\begin{table}[t]

    \centering
    \footnotesize
    \begin{tabular}{@{}l@{\hspace{0.1cm}}cccc@{}}
        \toprule
        \multirow{2}{*}{\centering\makecell{Weight \\ Type}} & \multicolumn{4}{c}{\textit{OpenVLA} (\( >2.0 \, \text{TFLOPs} \))} \\
        \cmidrule(lr){2-5}
         & Memory & Latency & SpeedUp & Energy Saving \\
        \midrule
        BF16 & 15.2 GB & 955.2 ms & $1.0 \times$ & $1.0 \times$ \\
        INT8 & 7.9 GB & 573.6 ms & $1.6 \times$ & $1.7 \times$ \\
        INT4 & 4.0 GB & 374.7 ms & $2.5 \times$ & $2.5 \times$ \\
        \bottomrule
    \end{tabular}
\vspace{-0 cm}
    \caption{Performance Metrics for \textit{OpenVLA} on GPU (NVIDIA Jetson AGX Orin): Memory, latency, and energy saving across various precision formats.}
    \label{tab:OpenVLA_Performance}
\vspace{-0 cm}
\end{table}

\section{Conclusion}

This paper proposes a quantization framework for IL-based models that enhances robustness against low-bit precision errors, ensuring efficiency on resource-limited hardware. Evaluations on robot manipulation and self-driving models show superior speedups and energy savings on real CPU and GPU, closely preserving full-precision accuracy and demonstrating practical deployment potential.

{
    \small
    \bibliographystyle{ieeenat_fullname}
    \bibliography{main}
}

\appendix
\clearpage
\section{Appendix}
\label{sec:appendix}

\subsection{Experiments Details}
\label{sec:appendix_exp}
\subsubsection{Benchmark Details}

\begin{figure}[h]
\centering
{\includegraphics[width=1\columnwidth]{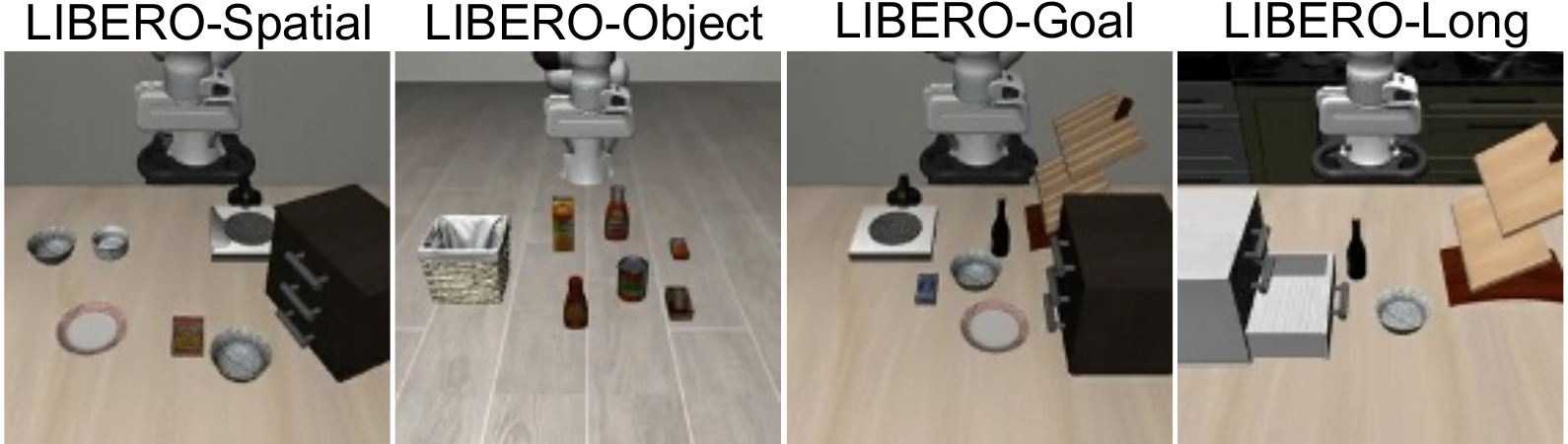}}
\caption{LIBERO benchmark.}
\label{fig:libero_bench}
\vspace{-0.3 cm}
\end{figure}

\textbf{Robot Manipulation:} We employed the LIBERO benchmark~\cite{liu2024libero} to assess the efficacy of QAIL and QBC within the realm of robot manipulation. The LIBERO benchmark includes four distinct task suites, each designed to facilitate life-long learning studies. Our research focused on implementing quantization techniques during the imitation learning process across these suites, subsequently evaluating the robustness and performance of the resulting quantized policies. Each suite comprises 10 distinct tasks accompanied by 50 human teleoperation demonstrations. LIBERO-Spatial evaluates spatial relationship understanding through varied object layouts; LIBERO-Object tests policy performance across different objects within identical layouts; LIBERO-Goal examines task-oriented behavior under consistent object and layout conditions with varied goals; and LIBERO-Long involves comprehensive long-horizon tasks that incorporate diverse objects, layouts, and objectives. To illustrate, the following are examples of tasks from each suite, with corresponding visualizations provided in Fig.~\ref{fig:libero_bench}:
\begin{itemize}
    \item \textbf{LIBERO-Spatial:} ``Pick up the black bowl \textbf{next to the ramekin} and place it on the plate."
    \item \textbf{LIBERO-Object:} ``Pick up \textbf{the butter} and place it in the basket."
    \item \textbf{LIBERO-Goal:} ``Open the middle drawer of the cabinet."
    \item \textbf{LIBERO-Long:} ``Put the black bowl in the bottom drawer of the cabinet and close it."
\end{itemize}

In our experiments, we used datasets that were specifically modified for compatibility with the OpenVLA~\cite{kim24openvla} framework, which included enhancements such as high-resolution image processing, image rotation, and the exclusion of unsuccessful demonstrations. We conducted 500 experiments for each task suite.
\begin{figure*}
    \centering
    \includegraphics[width=0.8\linewidth]{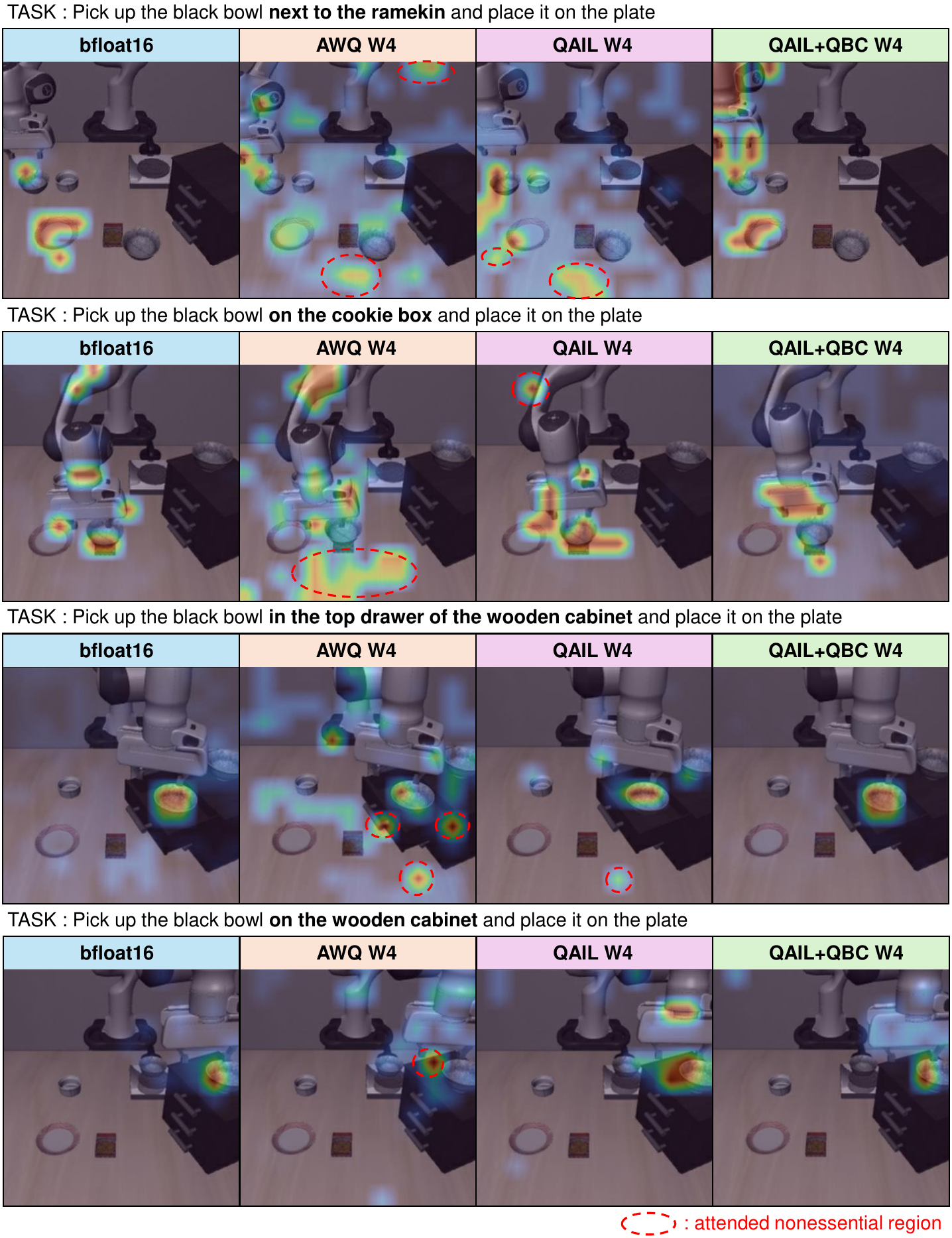}
\vspace{-0.25 cm}
   \caption{Comparison of attention visualization for tasks successfully completed on the LIBERO-Spatial benchmark.}
\vspace{-0.15 cm}
\label{fig:attention_sup}
\end{figure*}

{\setlength{\parindent}{0pt}
\textbf{Autonomus Driving:}
For evaluation, we utilized the NoCrash benchmark~\cite{codevilla2019exploring}. This benchmark evaluates the generalization capabilities from Town 1, characterized by its European town setup with single-lane roads and T-junctions, to Town 2, noted for its smaller scale and distinct textural differences. The benchmark includes three traffic density levels: empty, regular, and dense. These levels define the number of pedestrians and vehicles present in each map scenario. We conducted our performance evaluations under the NoCrash-dense setting, using metrics such as the success rate proposed by NoCrash and the driving score from the CARLA LeaderBoard~\cite{leaderboard}, along with rewards derived from CARLA~\cite{dosovitskiy2017carla}. The success rate is the proportion of routes completed without collisions or blockages, while the driving score is calculated based on penalties for infractions. Our infraction analysis measures occurrences such as pedestrian and vehicle collisions and red light violations per kilometer.
}

{\setlength{\parindent}{0pt}
\textbf{Physics Simulation Tasks:}
Performance evaluation in physics simulation tasks involved measuring the average return values for each task within the DeepMind Control Suite~\cite{tassa2018deepmind}. Detailed descriptions of each task are as follows:
\begin{itemize}
    \item \textbf{Cartpole Balance}: The task requires controlling a cart to keep a pole upright by moving the cart along a track, focusing on balance and stability.
    \item \textbf{Walker Stand}: This task involves maintaining an upright posture for a bipedal walker robot without it undertaking any additional locomotion.
    \item \textbf{Hopper Stand}: A one-legged robot must remain stable and upright, testing the agent's ability to balance a dynamically unstable object.
    \item \textbf{Cheetah Run}: The goal is to maximize the forward velocity of a quadrupedal cheetah robot, emphasizing speed control and efficient limb coordination.
    \item \textbf{Finger Spin}: The agent must control a robotic finger to spin an unactuated body continuously, assessing precision and control consistency.
    \item \textbf{Humanoid Stand}: The task requires a humanoid robot to maintain a standing position, testing balance and stability in a complex robot with many degrees of freedom.
    \item \textbf{Humanoid Walk}: Extending the humanoid stand task, this requires the robot to walk at a specified speed, focusing on the coordination of bipedal locomotion.
\end{itemize}
}

\subsubsection{Hyperparameter Setting}
\textbf{Robot Manipulation:}
For robot manipulation tasks, we utilized channel-wise quantization for weights. Both QAIL and QAIL+QBC models commence training from weights initialized using PTQ (AWQ~\cite{lin2023awq}). QLORA~\cite{dettmers2024qlora} (r=32) is employed to freeze the quantized weights, allowing updates solely to the adaptor, which comprises 110M parameters. The discrepancy metric \(D\) used is the average of the L2 distances. The weighted hyperparameter \(\lambda\) for QBC was set to 1. For wQBC, the hyper-parameter \(\beta\) is set at 2. Training proceeds wih a learning rate of 5e-4 for a total 50,000 steps.

{\setlength{\parindent}{0pt}
\textbf{Autonomus Driving:}
In autonomous driving, tensor-wise quantization is employed for both weights and activations using LSQ~\cite{esser2020learned}. The discrepancy metric \(D\) utilized measures the average of the L2 distances of the policy network’s final logits. Data collection utilizes the FP policy on the CARLA, capturing 80 episodes. The quantized policy is trained over 15 epochs with a learning rate of 1e-4. The weighted hyperparameter \(\lambda\) for QBC is set at 1.
}

{\setlength{\parindent}{0pt}
\textbf{Physics Simulation Tasks:}
For physics simulation tasks, tensor-wise quantization for both weights and activations is achieved using LSQ. The discrepancy metric \(D\) used is the average of the L2 distances of the policy network’s final logits. Training uses demonstration data with a learning rate of 3e-4 and extends over 1,000,000 steps. The weighted hyperparameter \(\lambda\) for QBC is set at 1.
}

\subsection{Attention Map Analysis}
\label{sec:appendix_map}
Additional visualizations are provided in Fig.~\ref{fig:attention_sup} for Sec.~\ref{sec:analysis}, illustrating a broad spectrum of tasks. The full-precision (FP) policy consistently demonstrates precise focus on relevant task objects and their specific locations, particularly where interactions with the robot arm are likely to occur. In contrast, policies quantized using PTQ (AWQ) often misdirect attention towards irrelevant areas, frequently overlooking the critical zones necessary for successful task execution. QAIL represents an improvement, more accurately targeting relevant object areas, though it occasionally still attends to non-essential regions. The integration of QAIL+QBC significantly enhances this focus, motivating a closer alignment of the quantized policy's attention with that of the FP policy. This alignment contributes to actions that are more likely based on relevant and accurate situational awareness, reflecting the reasoning processes of the FP policy.

\subsection{Quantization-Aware RL + QBC}
\label{sec:appendix_qarl}
\subsubsection{Method}

\begin{table*}[t]
\centering
\resizebox{1.6\columnwidth}{!}{%
\begin{tabular}{l|c|cccc|cccc|cccc}
\Xhline{2\arrayrulewidth}
\multirow{2}{*}{Method} & \multirow{2}{*}{Bit-width} & \multicolumn{4}{c|}{Suc. Rate \% $\uparrow$} & \multicolumn{4}{c|}{Dri. Score \% $\uparrow$} & \multicolumn{4}{c}{Reward $\uparrow$} \\ \cline{3-14} 
 &  & tt & tn & nt & nn & tt & tn & nt & nn & tt & tn & nt & nn \\ \Xhline{2\arrayrulewidth}
RL & FP & 92 & 90 & 82 & 84 & 96 & 94 & 89 & 91 & 2327 & 2219 & 1945 & 1959 \\ \midrule
\multirow{2}{*}{QARL} & W8A8 & 88 & 88 & 74 & 76 & 92 & 91 & 85 & 87 & 2107 & 2144 & 1517 & 1694 \\  
 & W4A4 & \textbf{94} & \textbf{90} & 68 & 76 & \textbf{95} & \textbf{94} & 80 & \textbf{84} & 2176 & 1998 & 1402 & 1214 \\ 
QARL+QBC & W4A4 & 92 & \textbf{90} & \textbf{84} & \textbf{78} & \textbf{95} & 92 & \textbf{90} & \textbf{84} & \textbf{2363} & \textbf{2355} & \textbf{1808} & \textbf{1739} \\ \Xhline{2\arrayrulewidth}
\end{tabular}%
}
\caption{Comparison of success rate, driving score, and reward on the NoCrash-dense benchmark with \textit{Roach} (tt: train-town \& train-weather, tn: train-town \& new-weather, nt: new-town \& train-weather, nn: new-town \& new-weather).}
\label{tab:roach_main}
\end{table*}

\begin{figure}[t]
{\includegraphics[width=\columnwidth]{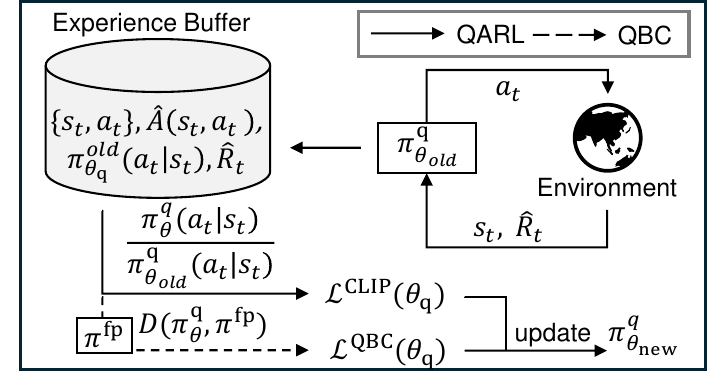}}
\caption{Overview of QARL+QBC with PPO }
\label{fig:qarl}
\end{figure}

Our original objective was to enhance quantization performance across various imitation learning (IL) domains, where the integration of Quantization-robust Behavior Cloning (QBC) proved successful. To explore the generality of the QBC concept in models targeting reward optimization, we expanded our investigation to include experiments applying QBC within reinforcement learning models. This exploration led to the development of the Quantization-Aware Reinforcement Learning (QARL) framework, where our primary objective is to ensure that a quantized policy, $\pi^{q}$, guarantees rewards even when operating under reduced precision in weights and activations. This setup is ideally suited for deployment on resource-constrained devices. The methodology modifies the reinforcement learning process to incorporate quantization during data collection and policy updates, maintaining performance stability despite the constraints of reduced precision.
To achieve this, we structure the QARL process into three key phases: data collection, policy optimization, and the integration of QBC, as illustrated in Fig.~\ref{fig:qarl}.

{\setlength{\parindent}{0pt}
\textbf{1. Data Collection:} 
The agent interacts with the environment under the guidance of the quantized policy $\pi^{q}_{\theta_{\text{old}}}$. During this process, key information about states, actions, and rewards is recorded at each timestep and stored in an experience buffer for future reinforcement learning updates.
}

{\setlength{\parindent}{0pt}
\textbf{2. Policy Optimization:} To optimize the policy, we employ Proximal Policy Optimization (PPO)~\cite{schulman2017proximal}, which utilizes a clipped objective function to carefully manage the extent of policy updates. This approach ensures that modifications remain within a permissible range, preventing any degradation in the policy’s performance. The objective function is defined as follows:
\begin{equation}
\label{eq:ppo_objective}
\mathcal{L}^{CLIP}(\theta) = \mathbb{E}_t \left[ \min \left( r_t(\theta) \hat{A}_t, \text{clip}(r_t(\theta), 1 - \epsilon, 1 + \epsilon) \hat{A}_t \right) \right],
\end{equation}
where \(r_t(\theta) = \frac{\pi_\theta(a_t | s_t)}{\pi_{\theta_{\text{old}}}(a_t | s_t)}\) is the ratio of new to old policy probabilities, $\hat{A}_t$ is the advantage estimation, and $\epsilon$ is the clipping threshold.
}

{\setlength{\parindent}{0pt}
\textbf{3. Integration of QBC:} To enhance $\pi^{q}$’s performance further and ensure stability in complex scenarios, we integrate the QBC as introduced in equation~\ref{eq:qbc_objective} of Sec.~\ref{sec:qbc}. QBC aids in aligning the decision-making of the quantized policy with that of a pre-trained full-precision (FP) policy, thereby enhancing decision accuracy. The combined loss function integrates $\mathcal{L}^{CLIP}$ from PPO with $\mathcal{L}^{QBC}$ from QBC:
\begin{equation}
\label{eq:total_objective_qarl}
\mathcal{L}^{\text{total}}(\theta) = \mathcal{L}^{CLIP}(\theta) + \lambda \mathcal{L}^{QBC}(\theta),
\end{equation}
where $\lambda$ is a hyperparameter that balances the influence of QBC.
}
This dual approach not only preserves the intrinsic advantages of the FP policy but also extends these benefits to the quantized model, ensuring its effectiveness in diverse and challenging operational environments.

\subsubsection{Experiment reuslts}
Table~\ref{tab:roach_main} demonstrates that applying QARL alone results in a notable decrease in success rate, driving score, and reward in new-town scenarios compared to the full-precision (FP) policy. However, integrating QBC alongside QARL significantly improves performance, bringing it closer to the levels achieved by the FP policy.

Table~\ref{tab:qarl_collision_comparison} presents a detailed comparison of infractions observed with the quantized policy. When QARL is applied in the W4A4 configuration without QBC, the model exhibits increased infractions, including pedestrian collisions, vehicle collisions, and red light violations. In contrast, incorporating QBC with QARL eliminates pedestrian collisions and reduces both vehicle collisions and red light infractions. 

To assess the generalizability of our proposed algorithms, we extend the evaluation beyond autonomous driving benchmarks to classical reinforcement learning tasks in the DeepMind Control Suite. The results, summarized in Table~\ref{tab:qarl_comparison}, allow for a comparison of our quantization techniques, QARL+QBC, with established approaches such as QuaRL~\cite{faust2022quarl}. These experiments further confirm the broad applicability of QBC across diverse learning paradigms, underscoring its pivotal role in addressing challenges associated with policy quantization and ensuring robust performance across varying scenarios.

\begin{table}[t]
\centering
\resizebox{\columnwidth}{!}{%
\begin{tabular}{l|c|c|c|c}
\Xhline{2\arrayrulewidth}
\multirow{3}{*}{Method} & \multirow{3}{*}{Bit-width}  & \multicolumn{3}{c}{Metrics $\downarrow$} \\ \cline{3-5}
 &   & Collision & Collision & Red light \\ 
 &   & pedestrian & vehicle & infraction \\ \Xhline{2\arrayrulewidth}
RL & FP  & 0 & 0.27 & 0.15 \\ \midrule
QARL  & W4A4 & 0.23 & 1.11 & 0.46 \\ 
QARL+QBC & W4A4  & 0 & \textbf{0.23} & \textbf{0.24} \\  \Xhline{2\arrayrulewidth}
\end{tabular}%
}

\caption{Comparison of collision and red light infraction metrics in a new-town \& train-weather scenario with \textit{Roach} on the NoCrash-dense benchmark.}
\label{tab:qarl_collision_comparison}

\end{table}

\begin{figure*}[t]
{\includegraphics[width=\textwidth]{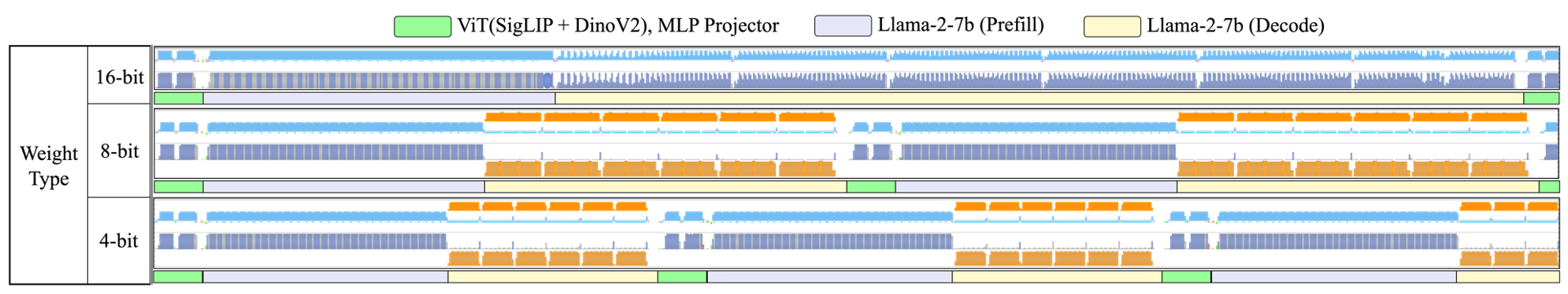}}
\caption{Hardware execution timeline comparison of 16-bit, 8-bit, and 4-bit weight datatypes for \textit{OpenVLA}, measured using \textit{NVIDIA Nsight Systems} over a total duration of 1,000 ms. The visualization highlights differences in execution patterns and latency across the three weight datatypes. Each stage is distinguished by different colors.}
\label{fig:nsys_report}
\end{figure*}
\subsection{Implementation Details}
\label{sec:appendix_imp}
\subsubsection{Detailed settings }
\begin{table}[t]
\centering
\resizebox{0.9\columnwidth}{!}{%
\begin{tabular}{l|c||c|c}
\Xhline{2\arrayrulewidth}
\multirow{2}{*}{TASK} &FP32 & \multicolumn{2}{c}{INT4} \\ \cline{2-4} 
 & RL & QuaRL\cite{faust2022quarl} & QARL+QBC  \\ \hline
Cartpole Balance & 981 & 519 & \textbf{845} \\ \hline
Walker Stand & 925 & 504 & \textbf{818}  \\ \hline
Hopper Stand & 833 & 425 & \textbf{754}  \\ \hline
Cheetah Run & 725 & 362 & \textbf{692}  \\ \hline
Finger Spin & 815 & 432 & \textbf{725}  \\ \hline
Humanoid Stand & 871 & 455 & \textbf{685}  \\ \hline
Humanoid Walk & 624 & 326 & \textbf{549}  \\ \Xhline{2\arrayrulewidth}
\end{tabular}%
}
\caption{Comparison of average return for each task by quantization method using D4PG~\cite{barth2018distributed} on DeepMind Control Suite.}
\label{tab:qarl_comparison}
\end{table}
\begin{figure}[h]
{\includegraphics[width=\columnwidth]{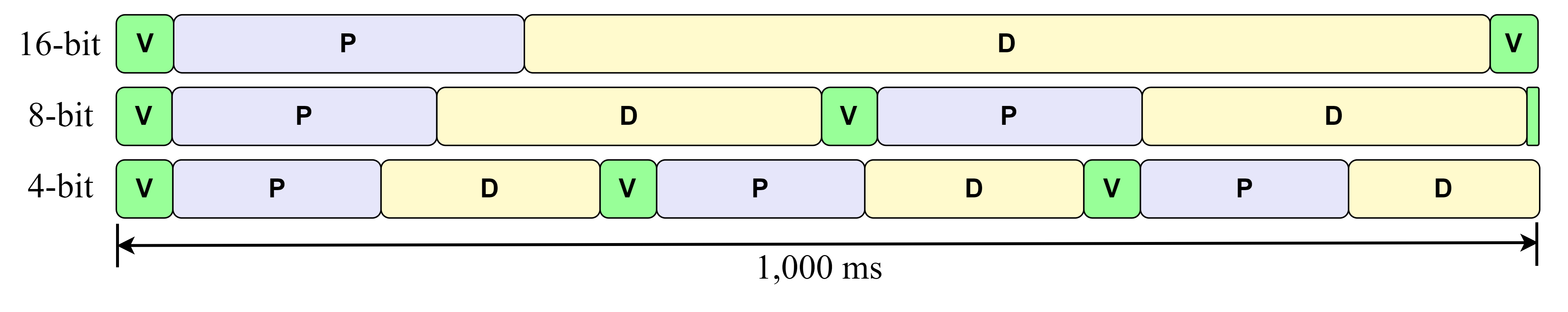}}
\caption{Timeline comparison of \textit{OpenVLA} for 16-bit, 8-bit, and 4-bit datatypes over a total duration of 1,000 ms. V represents the ViT+MLP projector, P denotes the prefill operation, and D refers to the decode operation of the backbone LLM (Llama-2-7b).}
\label{fig:timeline_openvla}
\end{figure}

\textbf{Device Settings:} For our experimental setup, we utilized NVIDIA Jetson AGX Orin 64GB and RTX 2080Ti GPU. NVIDIA Jetson AGX Orin 64GB is equipped with a 12-core Arm Cortex-A78AE CPU, an NVIDIA Ampere architecture GPU. The device runs on Ubuntu 20.04 64-bit LTS OS with GNU gcc/g++ version 9.3.0 with a 30W power mode. For each experiments, we employ the following devices:
\begin{itemize}
    \item Autonomous Driving (\textit{CILRS}, Weight-Activation Quantization):
    \begin{itemize}
        \item CPU (FP32, FP16, W8A8): NVIDIA Jetson AGX Orin 64GB (ARM Cortex-A78AE)
        \item GPU (FP32, W8A8, W4A4): NVIDIA RTX 2080 Ti
    \end{itemize}
    \item Vision-Language Action (\textit{OpenVLA}, Weight Quantization):
    \begin{itemize}
        \item GPU (16-bit, 8-bit, 4-bit): NVIDIA Jetson AGX Orin (2048-core NVIDIA Ampere GPU)
    \end{itemize}
\end{itemize} A comprehensive summary of the hardware specifications employed in our experiments is provided in Table~\ref{tab:hardware_specs_updated}.

\begin{table}[t]
    \renewcommand{\arraystretch}{1.2}
    \centering
    \footnotesize
    \begin{tabular}{@{}c@{\hspace{0.1cm}}cc@{}}
        \toprule
        \multirow{2}{*}{\centering\makecell{\textbf{Specification}}} & \textbf{NVIDIA Jetson AGX Orin} & \textbf{Portable Workstation} \\
        \cmidrule(lr){2-3}
         & \textbf{CPU/GPU} & \textbf{GPU} \\
        \midrule
        \makecell{\textbf{Computing} \\
        \textbf{Architecture}} 
        & \makecell{ARM Cortex-A78AE \\ 12 Cores, 2.2GHz \\ 2048-core NVIDIA Ampere \\ GPU with 64 Tensor Cores} 
        & \makecell{GeForce RTX 2080 Ti} \\
        \midrule
        \textbf{Cache (L1/L2)} 
        & \makecell{CPU: 64KB/256KB \\ GPU: 3MB/4MB} 
        & \makecell{GPU: 64KB/5.5MB} \\
        \midrule
        \textbf{Memory} 
        & 64GB LPDDR5 SDRAM 
        & 11GB GDDR6 \\
        \midrule
        \textbf{ISA} 
        & ARM v8.2-A (64 bit) / CUDA 12.0 
        & CUDA 12.0 \\
        \bottomrule
    \end{tabular}
    \caption{Hardware Specifications of NVIDIA Jetson AGX Orin and Portable Workstation.}
    \label{tab:hardware_specs_updated}
\end{table}

\textbf{Energy Consumption Measurement:} To measure energy consumption in our experiments, we employ the jetson-stats library, which is specifically designed for use with NVIDIA Jetson devices. This library leverages the capabilities of the Triple Channel Voltage/Current Monitor (Texas Instrument INA3221) integrated into NVIDIA Jetson devices~\cite{tegrastats_toolkit}. The INA3221 sensor provides detailed measurements of voltage, current, and power consumption for various power rails on the device, allowing for precise monitoring and analysis of the on board power usage.

\subsubsection{Analysis}

\begin{figure}[t]
{\includegraphics[width=\columnwidth]{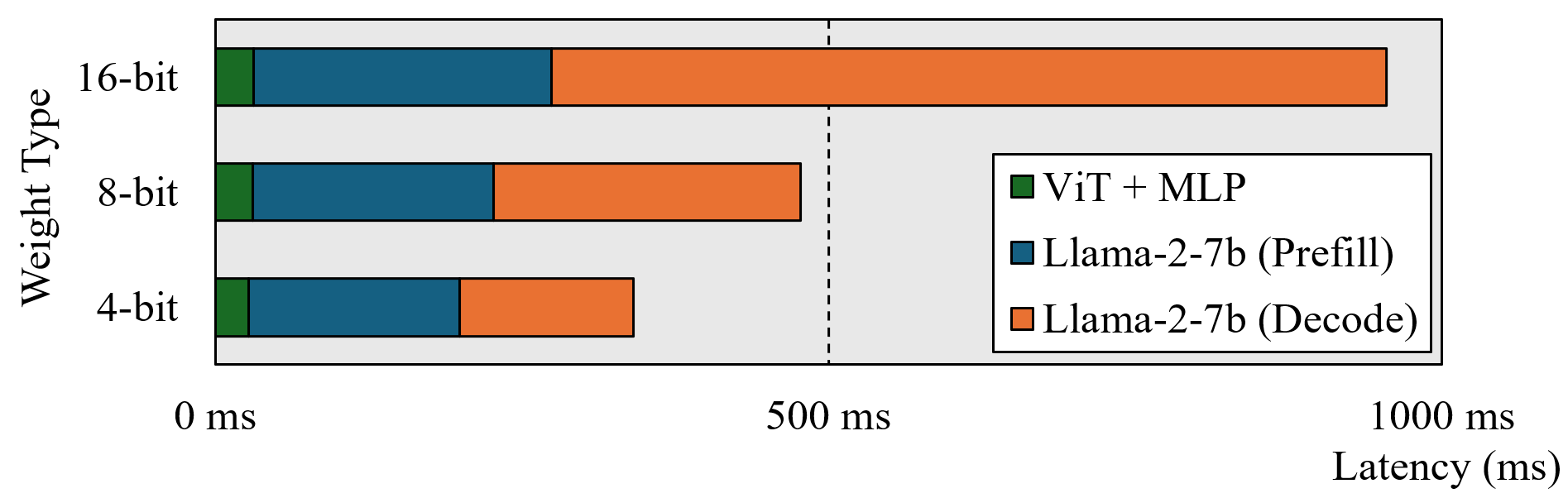}}
\caption{Latency breakdown of \textit{OpenVLA} for 16-bit, 8-bit, and 4-bit weights within a single step, including ViT+MLP, Llama-2-7b (Prefill), and Llama-2-7b (Decode) stages.}
\label{fig:latency_breakdown}
\end{figure}

We provide a detailed analysis and breakdown of the inference computation time of the \textit{OpenVLA} model. The inference process is divided into four main stages: (1) ViT (SigLIP + DINOv2), (2) MLP Projector, (3) Backbone LLM (Prefill), and (4) Backbone LLM (Decode). 

\textbf{NVIDIA Nsight Systems Hardware Execution Report:} NVIDIA Nsight Systems is a comprehensive performance analysis tool that captures and visualizes detailed hardware execution traces. In Figure~\ref{fig:nsys_report}, we present the analysis results of the hardware execution trace recorded for weight types of 16-bit, 8-bit, and 4-bit. 

\textbf{Timeline Comparison \& Breakdown:} Figure~\ref{fig:timeline_openvla} shows a timeline comparison for each weight type based on the actual time ratio, illustrating the proportion of time reduction achieved with different weight types. Figure~\ref{fig:latency_breakdown} presents the latency breakdown for each stage and analyzes how reducing the weight type affects the execution time. As shown in our experimental results, the latency of the decode operation in the backbone LLM is reduced the most, achieving up to a 2.5$\times$ speedup in overall execution time.

These experimental results confirm that the decode operation in the inference stage of the \textit{OpenVLA} model exhibits memory-bound characteristics, as demonstrated by our data. Weight quantization reduces the amount of data that needs to be transferred between memory and processing units, thereby alleviating memory bandwidth limitations and enhancing overall execution speed. By employing weight quantization techniques, we alleviate these memory-bound limitations, resulting in actual speedups in hardware execution time on Vision-Language Action model.

\label{sec:exp_details}

\end{document}